\documentclass[conference,a4paper]{IEEEtran}
\IEEEoverridecommandlockouts

\usepackage[hidelinks]{hyperref}
\usepackage[cmex10]{amsmath}
\usepackage{amssymb,amsfonts}
\usepackage{dblfloatfix}

\usepackage[ruled,vlined]{algorithm2e}
\usepackage{graphicx}
\graphicspath{{Figures/PDF/}{Figures/PNG/}}

\usepackage{booktabs}
\usepackage{siunitx}
\usepackage[numbers,compress]{natbib}
\usepackage{texnames}
\usepackage{bm,bbm}
\usepackage{orcidlink}

\begin{document}

\bstctlcite{IEEEexample:BSTcontrol}

\title{\uppercase{Enhancing SAR Object Detection with Self-Supervised Pre-training on Masked Auto-Encoders}}

\author{	\IEEEauthorblockN{Xinyang Pu\orcidlink{0009-0002-0627-4603}}
	\IEEEauthorblockA{
    \textit{Key Lab for Information Science} \\
        \textit{of Electromagnetic Waves (MoE)}, \\
        Fudan University,
        Shanghai, China, 200433 \\
        xypu21@m.fudan.edu.cn}
    
	\and
	\IEEEauthorblockN{Feng Xu \orcidlink{0000-0002-7015-1467}}
	\IEEEauthorblockA{
    \textit{Key Lab for Information Science} \\
       \textit{ of Electromagnetic Waves (MoE)}, \\
        Fudan University, 
        Shanghai, China, 200433 \\
        fengxu@fudan.edu.cn}
    
}


\maketitle
\begin{abstract}
	Supervised fine-tuning methods (SFT) perform great efficiency on artificial intelligence interpretation in SAR images, leveraging the powerful representation knowledge from pre-training models. Due to the lack of domain-specific pre-trained backbones in SAR images, the traditional strategies are loading the foundation pre-train models of natural scenes such as ImageNet, whose characteristics of images are extremely different from SAR images. This may hinder the model performance on downstream tasks when adopting SFT on small-scale annotated SAR data. In this paper, an self-supervised learning (SSL) method of masked image modeling based on Masked Auto-Encoders (MAE) is proposed to learn feature representations of SAR images during the pre-training process and benefit the object detection task in SAR images of SFT. The evaluation experiments on the large-scale SAR object detection benchmark named SARDet-100k verify that the proposed method captures proper latent representations of SAR images and improves the model generalization in downstream tasks by converting the pre-trained domain from natural scenes to SAR images through SSL. The proposed method achieves an improvement of 1.3 mAP on the SARDet-100k benchmark compared to only the SFT strategies.
\end{abstract}


\section{Introduction}
Supervised learning has been widely applied in the intelligent interpretation of synthetic aperture radar (SAR) images. Compared to training models from scratch, supervised fine-tuning (SFT) based on large-scale pre-trained models has demonstrated significant advantages. However, due to the lack of pre-trained backbones specifically designed for SAR images, conventional approaches often resort to utilizing pre-trained weights from large-scale natural image datasets, such as ImageNet, for fine-tuning. This practice, may hinder the performance of models on specific downstream tasks due to the domain gap between natural images and SAR images.

In recent years, the increasing availability of SAR data has gradually enabled the development of pre-trained vision models specifically for SAR images. Due to the unique characteristics of SAR images and the challenges associated with their annotation and interpretation, existing SAR datasets typically contain only limited labeled data, which is often task-specific and intended for supervised learning in downstream tasks. Supervised fine-tuning on such limited labeled SAR data for specific tasks prevents models from leveraging the representation learning capabilities of large-scale unlabeled SAR data for optimization. Furthermore, supervised fine-tuning method tends to focus on mapping SAR images to task-specific annotations, which negatively impacts the generalization capability of feature extraction of models.

The growing availability of SAR images, combined with the scarcity of supervised labels and the independent training of downstream tasks, has hindered the effective utilization of SAR data for latent feature representation and learning through pre-training techniques. The masked image modeling (MIM) approach \cite{MAE}, a mainstream technique for vision model pre-training, enhances the model's feature extraction capabilities by masking parts of the image, inputting visible regions, and learning to reconstruct the masked areas. This improves the generalization performance of pre-trained backbones in downstream SFT tasks. 

With the significant advancements in data availability and computation resources, MIM-based pre-training methods have been widely applied to build foundational models for natural scenes and optical remote sensing images, demonstrating effectiveness across various downstream tasks. SAR-JEPA \cite{SAR-JEPA} employs SSL methods for SAR image pre-training and validated their effectiveness in image classification tasks, while FG-MAE \cite{FG-MAE} optimizes the reconstruction objective of the Masked Auto-encoders (MAE) algorithm to enhance model performance in scene classification and landcover segmentation tasks. However, pre-training models for SAR images and evaluating their performance on fine-grained downstream tasks, such as object detection, remain in the early stages. 

\begin{figure*}[h]
	\centering
	\includegraphics[width=\linewidth]{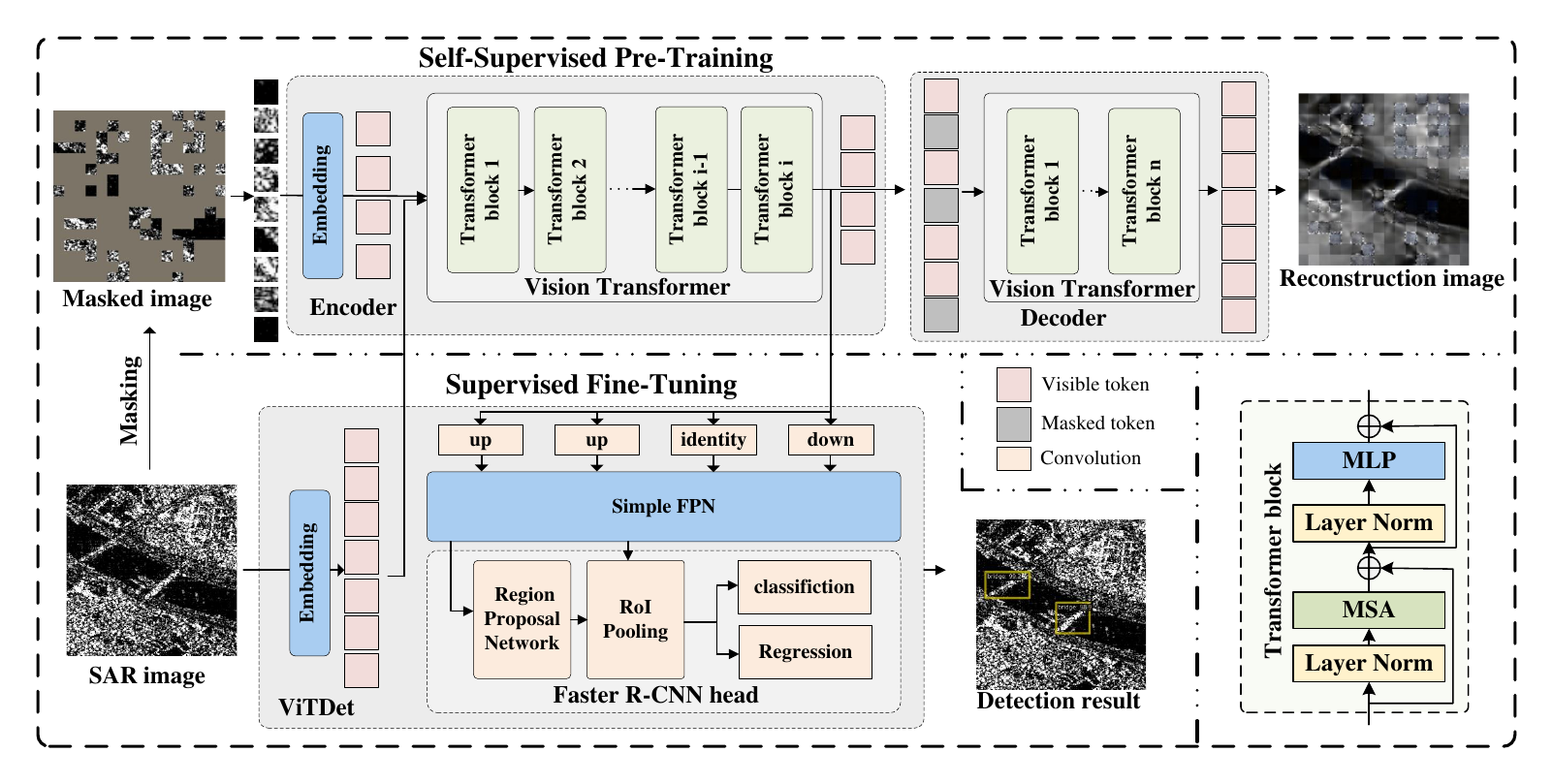}
	\caption{The architecture of the proposed method, including self-supervised pre-training and supervised fine-tuning.}
\end{figure*}

In this paper, the proposed method employs self-supervised learning with Masked Auto-encoders to reconstruct SAR images, enabling the model to learn robust feature representations, establish an effective pre-trained backbone, and improve performance in downstream supervised fine-tuning tasks. The evaluation experiments are conducted on the large-scale SAR object detection benchmark named SARDet-100k \cite{sardet}, which integrates multiple public SAR datasets for the object detection task. A pre-trained backbone of Vision Transformer \cite{VIT} for SAR images is built and employed to later supervised fine-tuning process. The proposed method achieves an improvement of 1.3 mAP from 53.6 to 54.9 in the detection performance on SARDet-100k validation set, verifying the efficiency of the pre-training approach based on self-supervised learning. 

\section{METHODOLOGY}
This section illustrates the proposed method in the following order: (a) self-supervised learning of pre-training based on Masked Auto-encoders, (b) supervised fine-tuning on object detection. The Vision Transformer encoder of MAE in the first step is employed as the pre-training backbone of ViT-Det \cite{ViTDet} on the second step of supervised fine-tuning. Figure 1 presents the entire architecture of the proposed method.

\subsection{SSL Pre-training based on Mask Auto-encoder}
Masked Auto-encoders \cite{MAE} have an asymmetric framework which consists of an encoder and a decoder both following Vision Transformer \cite{VIT} structure, shown in the upper part of Figure 1. The encoder extracts image patches to latent representation, and decoder maps the feature back to the original image. In this way, the MAE encoder learns proper feature extraction ability and improves the generalization of the model.

The input image \( x \in \mathbb{R}^{H \times W \times C} \) is divided into non-overlapping patches size of $p$, obtaining  \( N = \frac{H \times W}{p^2} \) image patches \( x_p \in \mathbb{R}^{N \times p \times p \times C} \), where \( (H, W) \) is the size of the input image, and \( C \) is the number of channels. A subset of patches \( x_m \in \mathbb{R}^{Nr \times p \times p \times C} \) is randomly sampled and masked according to a specified mask ratio $r\in [0, 1] $. The remaining unmasked patches \( x_v \in \mathbb{R}^{N(1-r) \times p \times p \times C} \), referred to as visible patches, are mapped into visible tokens through an embedding layer $f_p : \mathbb{R}^{p \times p \times C} \to \mathbb{R}^D $, where $D$ is the output embedding dimension. All visible tokens \( t_v \in \mathbb{R}^{N(1-r) \times  D} \) are concatenated with a cls token \( t_{cls} \in \mathbb{R}^{1 \times D} \), and then 2D positional embedding \( {PE}_{2D} \in \mathbb{R}^{[N(1-r)+1]\times D} \)are incorporated, after which the resulting sequence is input into the MAE encoder of the Vision Transformer architecture. 

The encoder, composed of a series of transformer blocks, extracts features from the visible tokens to produce latent representations. The output features of the encoder are concatenated with mask tokens and passed as input to a lightweight decoder. 
A high masking ratio $r$ (e.g., 75$\%$) is employed, where 75$\%$ of the patches are randomly masked, leaving only 25$\%$ of the patches as visible tokens input to the encoder. The lightweight decoder reconstructs the full set of tokens. This high masking ratio allows MAE to efficiently pre-train on large-scale image datasets while significantly reducing computational costs.

The decoder, also consisting of several ViT blocks with position embeddings, reconstructs the original image by predicting all patches, including both visible and masked ones. The reconstruction objective of MAE is to predict the pixel values of the masked patches \( x_m\), and the model is trained by minimizing the mean squared error (MSE) loss between the original pixel values of the masked patches \( x_m\) and the predicted ones \( x_{pred} \in \mathbb{R}^{Nr \times p \times p \times C} \). The loss of mean squared error is formulated as:

\begin{equation}
	{Loss}_{MSE} = \frac{1}{Nr} \sum_{i=1}^{Nr} ({x_{pred}}_i - {x_m}_i)^2,
\end{equation}

Since the encoder is responsible for image representation and feature extraction, it accounts for the majority of the model parameters of MAE, while the lightweight decoder is utilized for pixel reconstruction. After the pre-training phase, the decoder is discarded, and only the encoder is applied in the subsequent supervised fine-tuning stage.

\subsection{Supervised fine-tuning for object detection}
The MAE encoder adopts a Vision Transformer (ViT) architecture, known for its strong feature extraction and generalization in natural language processing and computer vision. To address the multi-scale and fine-grained demands of object detection, a Feature Pyramid Network (FPN) and Region of Interest (RoI) head are integrated with the pre-trained ViT backbone, forming the ViTDet framework \cite{ViTDet} for efficient object detection.

The bottom section of Figure 1 illustrates the ViTDet architecture in detail. The Vision Transformer backbone contains a patch embedding layer and several transformer blocks, consistent with the pre-train encoder. Simple FPN is responsible for building hierarchical feature map from a plain backbone for multi-scale detection. RoI head follows the Faster R-CNN \cite{fasterRCNN}, contains a Region Proposal Network (RPN), an RoI pooling and a classifier for final bonding boxes and classification.

ViT backbone acts as a plain, non-hierarchical backbone, after the patch embedding and position embedding, series of transformer blocks alternates between multi-head self-attention (MSA) layers and MLP blocks, with LayerNorm (LN) applied prior to each block and residual connections added following each block, maintaining a single-scale feature map with same dimension. Simple FPN is applied to generate multi-scale feature pyramid and benefit the object detection task. Simple FPN captures the last feature map of ViT backbone with the scale of $\frac{1}{16}$ and employs four convolution, identity or deconvolution layers with different strides $(2,1,\frac{1}{2},\frac{1}{4})$ to obtain multi-scale feature maps. 

The detection head of the proposed method follows the Faster R-CNN architecture. In RPN, region proposals are generated by sliding a small network over the pyramid feature map. An \(n \times n\) spatial window of the feature map is processed and mapped to a lower-dimensional feature which is the input of the final classifier. An RoI pooling is paralleled to the RPN and also fed into the classifier.  The final classifier contains two sibling fully connected layers: a regression layer of bounding boxes and a classification layer.

\section{EXPERIMENTS}
According to the illustrated paradigm of the proposed method, two main experiments are conducted: self-supervised learning of pre-training based on Mask Autoencoder and supervised learning of object detection task. Following the evaluation paradigm of SSL pre-training algorithms, the pre-trained model performance of MAE is qualitatively visualized in image reconstruction task and quantitatively evaluated in downstream supervised fine-tuning tasks, verifying the effectiveness of the proposed algorithm.

\subsection{Dataset}
SARDet-100K \cite{sardet} is a large-scale SAR object detection dataset with 116,598 images and 245,653 instances across six categories: Aircraft, Ship, Car, Bridge, Tank, and Harbor. Compiled from 10 public datasets to ensure diversity and consistency, it includes $512 \times 512$ cropped images divided into 94,493 for training, 10,492 for validation, and 11,613 for testing. Designed to bridge domain gaps, it supports robust SAR object detection and offers a dataset comparable to COCO dataset \cite{COCO} (118K images), fostering the development of advanced SAR detection models.

\subsection{Self-supervised pre-training of MAE}
\subsubsection{Parameters of the Network Structure}
For SSL pre-training, no annotations are required. All SARDet-100k training images are resized to $256 \times 256$ and processed by MAE. The input image is divided into $16 \times 16$ patches, generating 256 patches. With a mask ratio of 0.75, only 64 visible patches are fed into the ViT-Base encoder, which has 12 Transformer blocks with a dimension of 768. The MAE decoder, consisting of 3 Transformer blocks with a dimension of 512, reconstructs the masked image using both visible and masked tokens.

\subsubsection{Training configuration}
Pre-training uses random flipping with a 50\% probability. Two NVIDIA A800 80GB GPUs are used with a batch size of 256 per GPU. The AdamW optimizer is configured with a learning rate scaled as \(1.5 \times 10^{-4} \times \frac{512}{256}\), \(\beta=(0.9, 0.95)\), and a weight decay of 0.05. Training spans 400 epochs, with a linear warm-up for 40 epochs followed by cosine annealing. Automatic LR scaling is applied for a base batch size of 512.

\subsubsection{Qualitative Visualization of Reconstruction Results}
The primary objective of MAE pre-training is to enhance the encoder's feature extraction, not the decoder's reconstruction. Evaluation follows the standard MAE paradigm, focusing on supervised fine-tuning (SFT) performance rather than reconstruction metrics. Reconstruction results on the validation set are shown in Figure 2.

\begin{figure}[htbp]
  \centering
  \begin{minipage}{0.19\linewidth}
    \centering
    \includegraphics[width=\linewidth]{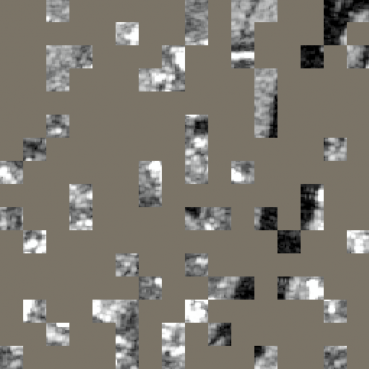}
  \end{minipage}
  \begin{minipage}{0.19\linewidth}
    \centering
    \includegraphics[width=\linewidth]{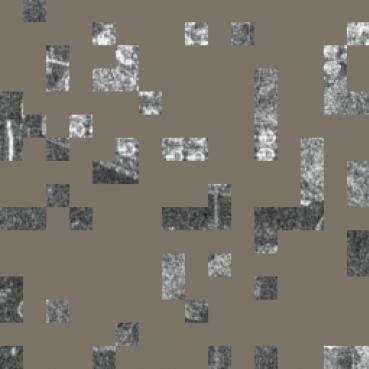}
  \end{minipage}
  \begin{minipage}{0.19\linewidth}
    \centering
    \includegraphics[width=\linewidth]{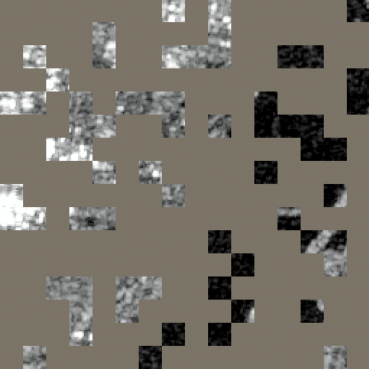}
  \end{minipage}
  \begin{minipage}{0.19\linewidth}
    \centering
    \includegraphics[width=\linewidth]{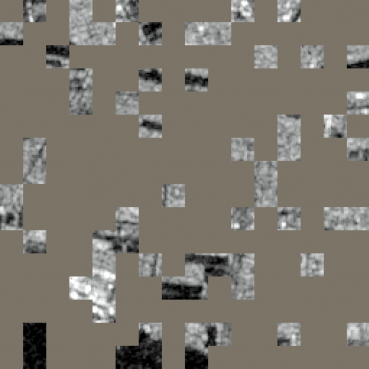}
  \end{minipage}
  \begin{minipage}{0.19\linewidth}
    \centering
    \includegraphics[width=\linewidth]{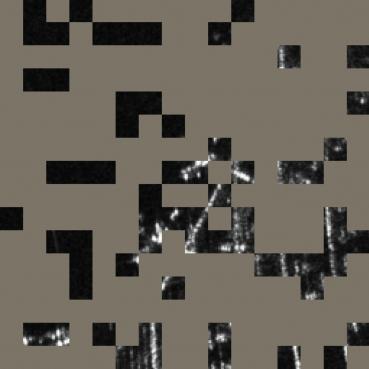}
  \end{minipage}

   \begin{minipage}{0.19\linewidth}
    \centering
    \includegraphics[width=\linewidth]{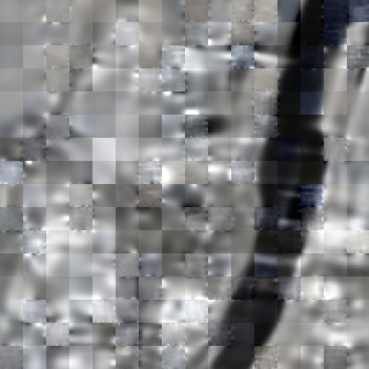}
  \end{minipage}
  \begin{minipage}{0.19\linewidth}
    \centering
    \includegraphics[width=\linewidth]{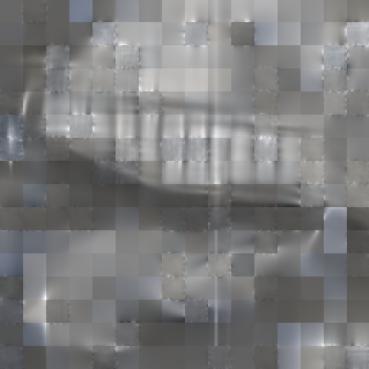}
  \end{minipage}
  \begin{minipage}{0.19\linewidth}
    \centering
    \includegraphics[width=\linewidth]{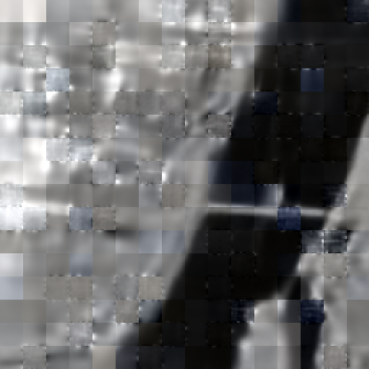}
  \end{minipage}
  \begin{minipage}{0.19\linewidth}
    \centering
    \includegraphics[width=\linewidth]{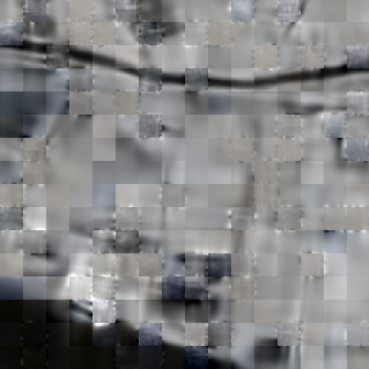}
  \end{minipage}
   \begin{minipage}{0.19\linewidth}
    \centering
    \includegraphics[width=\linewidth]{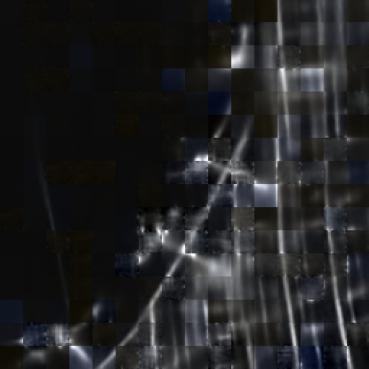}
  \end{minipage}

 \begin{minipage}{0.19\linewidth}
    \centering
    \includegraphics[width=\linewidth]{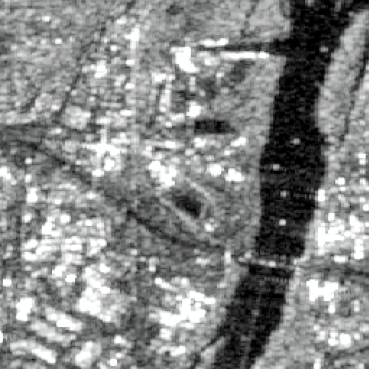}
  \end{minipage}
  \begin{minipage}{0.19\linewidth}
    \centering
    \includegraphics[width=\linewidth]{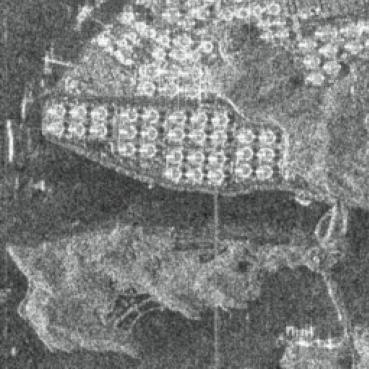}
  \end{minipage}
  \begin{minipage}{0.19\linewidth}
    \centering
    \includegraphics[width=\linewidth]{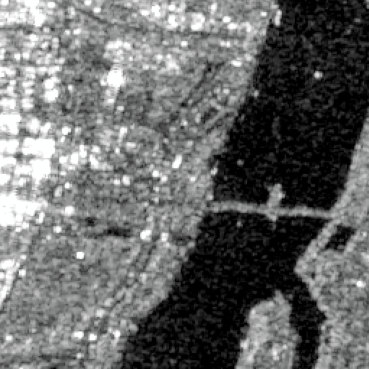}
  \end{minipage}
  \begin{minipage}{0.19\linewidth}
    \centering
    \includegraphics[width=\linewidth]{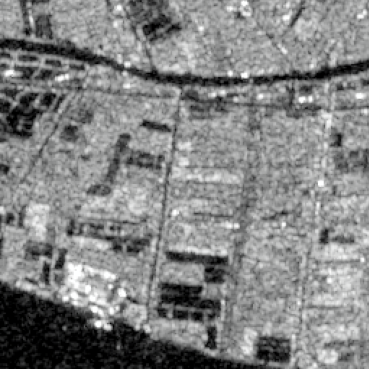}
  \end{minipage}
    \begin{minipage}{0.19\linewidth}
    \centering
    \includegraphics[width=\linewidth]{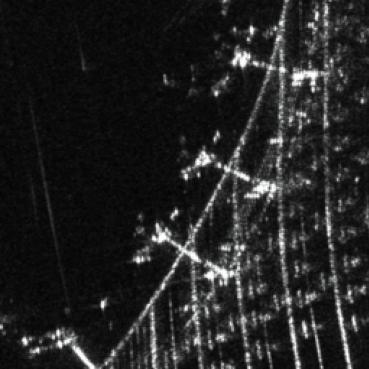}
  \end{minipage}

  \caption{ Reconstruction results of the MAE model on
the validation set. Row 1: Masked images. Row 2: Reconstructed images. Row 3: Ground truth. }
  \label{fig:main}
\end{figure}

\subsection{Supervised fine-tuning of object detection}
To evaluate the efficiency of the pre-trained backbone, the detection algorithm's architecture is kept simple and basic to avoid introducing too many interferences. The Faster R-CNN head with ViT-Base backbone named ViT-Det is introduced in the fine-tuning experiments.
\subsubsection{Parameters of the Network Structure}

The Vision Transformer backbone retains the pre-trained MAE encoder structure for weight compatibility in fine-tuning. The detection framework follows Faster R-CNN, including a simple FPN with five feature levels (256 channels each) and a standard RoI head. Feature maps have 192, 384, 768, and 768 channels. The RPN uses anchor boxes with aspect ratios $[0.5, 1.0, 2.0]$ and scales of 8, spanning strides $[4, 8, 16, 32, 64]$. RoIAlign ($7 \times 7$) is employed for spatial feature extraction.

\subsubsection{Training configuration}
The input images are $512 \times 512$ with a RandomFlip probability of 0.5. The model is trained for 12 epochs on 4 NVIDIA GeForce RTX 4090 GPUs, with a batch size of 4 per GPU. One GPU is used for evaluation. The AdamW optimizer is applied with an initial learning rate of 0.0001, betas of (0.9, 0.999), and a weight decay of 0.05. The learning rate schedule includes a linear warm-up for 500 steps and multi-step decay at epochs 8 and 11 (factor 0.1).

\subsubsection{Evaluation results of comparison experiments}
The performance of the SFT model is assessed using standard COCO object detection metrics, including mAP, AP50, AP75, and mAP for small, medium, and large objects. Comparative experiments are conducted between the MAE pre-trained ViT-Base models on ImageNet and SARDet-100k datasets. Additionally, a baseline model trained from scratch is included for reference. The evaluation results of SARDet-100k validation set are presented in Table 1. 

\begin{table}[hbt]
	\centering
	\caption{Detection Performance comparison on different pre-training.}\label{tab:performance}
    \resizebox{\columnwidth}{!}{%
	\begin{tabular}{lcccccc}
		\toprule
		\textbf{Pre-train data} & \textbf{mAP} & \textbf{AP50} & \textbf{AP75} & \textbf{mAP\_s} & \textbf{mAP\_m} & \textbf{mAP\_l} \\
        \cmidrule(lr){1-1} \cmidrule(lr){2-2}\cmidrule(lr){3-3}\cmidrule(lr){4-4} \cmidrule(lr){5-5}\cmidrule(lr){6-6}\cmidrule(lr){7-7}
		- & 47.2 & 80.0 & 49.6 & 40.7 & 56.9 & 50.6 \\ 
		ImageNet & 53.6 & 85.7 & 58.4 & 47.9 & 63.9 & 57.9 \\ 
		SARDet-100k & \textbf{54.9} & \textbf{85.9} & \textbf{60.5} & \textbf{50.0} & \textbf{65.3} & \textbf{58.7} \\ 
		\bottomrule
	\end{tabular}%
    }
\end{table}

\begin{figure}[htbp]
  \centering

   \begin{minipage}{0.19\linewidth}
    \centering
    \includegraphics[width=\linewidth]{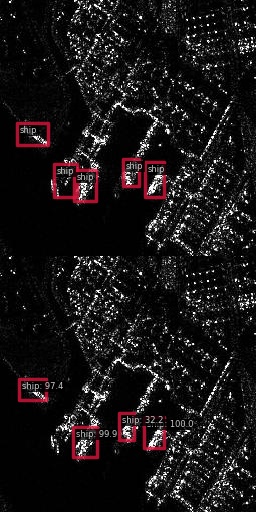}
  \end{minipage}
  \begin{minipage}{0.19\linewidth}
    \centering
    \includegraphics[width=\linewidth]{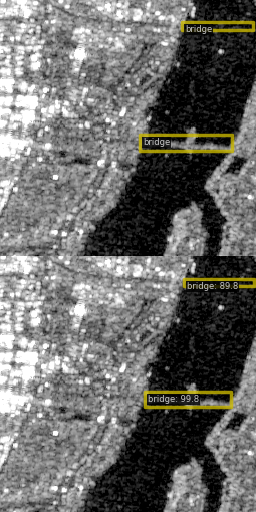}
  \end{minipage}
  \begin{minipage}{0.19\linewidth}
    \centering
    \includegraphics[width=\linewidth]{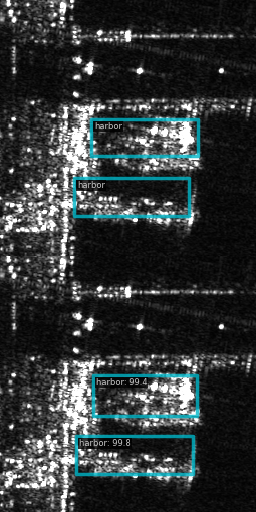}
  \end{minipage}
  \begin{minipage}{0.19\linewidth}
    \centering
    \includegraphics[width=\linewidth]{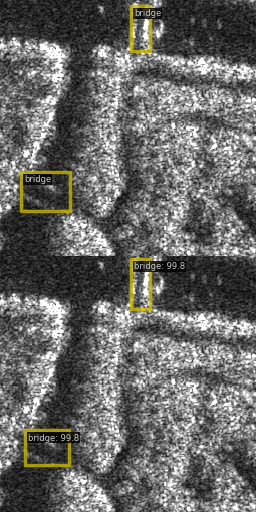}
  \end{minipage}
   \begin{minipage}{0.19\linewidth}
    \centering
    \includegraphics[width=\linewidth]{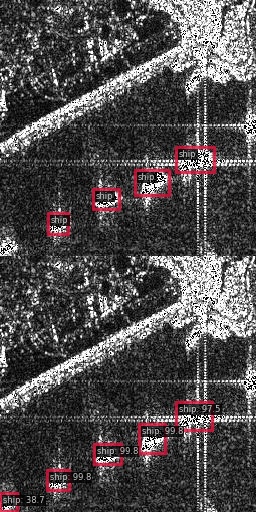}
  \end{minipage}

  \caption{Detection results of the ViTDet pre-trained on
SARDet-100k. Row 1: Ground truth. Row 2: Predicted result. Boxes in different colors indicate different categories. }
  \label{fig:main2}
\end{figure}

The proposed method exceeds the ImageNet pre-training on all the metrics, as well as training from scratch. Figure 3 presents the visualizations of the model detection results of the proposed method on SARDet-100k validation set. Such evaluation results reveal the importance of pre-training large-scale data of the consistent distribution for downstream fine-grained tasks. The model performance of training from scratch lag behind those of supervised fine-tuning, highlighting the importance of pre-training and the necessity of using SFT instead of initializing with random parameters of weights. 

The proposed method, which leverages self-supervised pre-training on SAR data, improves the performance of downstream tasks, demonstrating that, compared to conventional approaches that load ImageNet's pre-trained weights, the unique characteristics of SAR image domain make self-supervised pre-training on large-scale SAR data followed by SFT highly beneficial.

\section{Conclusion}
The proposed method investigates the use of self-supervised learning for pre-training on SAR images using the Mask Auto-encoders algorithm, aiming to enhance the generalization capability of pre-trained backbones. After SSL pre-training, supervised fine-tuning for object detection leads to significant improvements in model performance. Evaluation experiments on the large-scale SAR object detection benchmark, SARDet-100k, validate the necessity and effectiveness of the proposed approach in building a well pre-trained backbone tailored for the SAR domain, which effectively benefits fine-tuning downstream tasks.

\small
\bibliographystyle{IEEEtranN}
\bibliography{references}

\begin{thebibliography}{8}
\providecommand{\natexlab}[1]{#1}
\providecommand{\url}[1]{#1}
\csname url@samestyle\endcsname
\providecommand{\newblock}{\relax}
\providecommand{\bibinfo}[2]{#2}
\providecommand{\BIBentrySTDinterwordspacing}{\spaceskip=0pt\relax}
\providecommand{\BIBentryALTinterwordstretchfactor}{4}
\providecommand{\BIBentryALTinterwordspacing}{\spaceskip=\fontdimen2\font plus
\BIBentryALTinterwordstretchfactor\fontdimen3\font minus \fontdimen4\font\relax}
\providecommand{\BIBforeignlanguage}[2]{{%
\expandafter\ifx\csname l@#1\endcsname\relax
\typeout{** WARNING: IEEEtranN.bst: No hyphenation pattern has been}%
\typeout{** loaded for the language `#1'. Using the pattern for}%
\typeout{** the default language instead.}%
\else
\language=\csname l@#1\endcsname
\fi
#2}}
\providecommand{\BIBdecl}{\relax}
\BIBdecl
\renewcommand{\BIBentryALTinterwordstretchfactor}{4}

\bibitem[He et~al.(2022)He, Chen, Xie, Li, Doll\'ar, and Girshick]{MAE}
K.;He, X.;Chen, S.;Xie \emph{et~al.}, ``Masked autoencoders are scalable vision learners,'' in \emph{Proceedings of the IEEE/CVF Conference on Computer Vision and Pattern Recognition (CVPR)}, June 2022, pp. 16\,000--16\,009.

\bibitem[Li et~al.(2024{\natexlab{a}})Li, Yang, Liu, Hou, Li, Liu, Liu, and Liu]{SAR-JEPA}
W.;Li, W.;Yang, T.;Liu \emph{et~al.}, ``Predicting gradient is better: Exploring self-supervised learning for sar atr with a joint-embedding predictive architecture,'' \emph{ISPRS Journal of Photogrammetry and Remote Sensing}, vol. 218, pp. 326--338, 2024.

\bibitem[Wang et~al.(2025)Wang, Hernández, Albrecht, and Zhu]{FG-MAE}
Y.;Wang, H.~H.;Hernández, C.~M.;Albrecht \emph{et~al.}, ``Feature guided masked autoencoder for self-supervised learning in remote sensing,'' \emph{IEEE Journal of Selected Topics in Applied Earth Observations and Remote Sensing}, vol.~18, pp. 321--336, 2025.

\bibitem[Li et~al.(2024{\natexlab{b}})Li, Li, Li, Hou, Liu, Cheng, and Yang]{sardet}
Y.;Li, X.;Li, W.;Li \emph{et~al.}, ``Sardet-100k: Towards open-source benchmark and toolkit for large-scale sar object detection,'' \emph{arXiv preprint arXiv:2403.06534}, 2024.

\bibitem[Dosovitskiy et~al.(2021)Dosovitskiy, Beyer, Kolesnikov, Weissenborn, Zhai, Unterthiner, Dehghani, Minderer, Heigold, Gelly, Uszkoreit, and Houlsby]{VIT}
A.;Dosovitskiy, L.;Beyer, A.;Kolesnikov \emph{et~al.}, ``An image is worth 16x16 words: Transformers for image recognition at scale,'' in \emph{International Conference on Learning Representations}, 2021.

\bibitem[Li et~al.(2022)Li, Mao, Girshick, and He]{ViTDet}
Y.;Li, H.;Mao, R.;Girshick \emph{et~al.}, ``Exploring plain vision transformer backbones for object detection,'' in \emph{European conference on computer vision}.\hskip 1em plus 0.5em minus 0.4em\relax Springer, 2022, pp. 280--296.

\bibitem[Ren et~al.(2016)Ren, He, Girshick, and Sun]{fasterRCNN}
S.;Ren, K.;He, R.;Girshick \emph{et~al.}, ``Faster r-cnn: Towards real-time object detection with region proposal networks,'' \emph{IEEE transactions on pattern analysis and machine intelligence}, vol.~39, no.~6, pp. 1137--1149, 2016.

\bibitem[Lin et~al.(2014)Lin, Maire, Belongie, Hays, Perona, Ramanan, Doll{\'a}r, and Zitnick]{COCO}
T.-Y.;Lin, M.;Maire, S.;Belongie \emph{et~al.}, ``Microsoft coco: Common objects in context,'' in \emph{Computer Vision -- ECCV 2014}, D.;Fleet, T.;Pajdla, B.;Schiele \emph{et~al.}, Eds.\hskip 1em plus 0.5em minus 0.4em\relax Cham: Springer International Publishing, 2014, pp. 740--755.

\end{thebibliography}

\end{document}